\documentclass[lettersize,journal]{IEEEtran}
\usepackage{amsmath,amsfonts}
\usepackage{algorithmic}
\usepackage{algorithm}
\usepackage{array}
\usepackage{textcomp}
\usepackage{stfloats}
\usepackage{url}
\usepackage{verbatim}
\usepackage{graphicx}
\usepackage{cite}
\usepackage{caption,subcaption}
\usepackage{hyperref}
\usepackage{bbm}
\usepackage{dsfont}

\DeclareMathOperator*{\st}{s.t.}
\begin{document}

\title{Optimizing Value of Learning in Task-Oriented Federated Meta-Learning Systems}

\author{Bibo Wu\textsuperscript{\dag}, Fang Fang\textsuperscript{\dag, \ddag} and Xianbin Wang\textsuperscript{\dag}\\
	\textsuperscript{\dag}Department of Electrical and Computer Engineering, Western University, London, Canada\\
	\textsuperscript{\ddag}Department of Computer Science, Western University, London, Canada\\
	Emails: \{bwu293, fang.fang, xianbin.wang\}@uwo.ca
	}

\maketitle
\thispagestyle{empty} 
\begin{abstract}
Federated Learning (FL) has gained significant attention in recent years due to its distributed nature and privacy-preserving benefits. However, a key limitation of conventional FL is that it learns and distributes a common global model to all participants, which fails to provide customized solutions for diverse task requirements. Federated meta-learning (FML) offers a promising solution to this issue by enabling devices to fine-tune local models after receiving a shared meta-model from the server.
In this paper, we propose a task-oriented FML framework over non-orthogonal multiple access (NOMA) networks. A novel metric, termed value of learning (VoL), is introduced to assess the individual training needs across devices.
Moreover, a task-level weight (TLW) metric is defined based on task requirements and fairness considerations, guiding the prioritization of edge devices during FML training. The formulated problem—to maximize the sum of TLW-based VoL across devices—forms a non-convex mixed-integer non-linear programming (MINLP) challenge, addressed here using a parameterized deep Q-network (PDQN) algorithm to handle both discrete and continuous variables. Simulation results demonstrate that our approach significantly outperforms baseline schemes, underscoring the advantages of the proposed framework.

\end{abstract}


\section{Introduction}
In recent years, there has been a significant shift from centralized learning to federated learning (FL), driven by the rapid advancements in edge artificial intelligence \cite{FL_Mag}.
By only transmitting the local model parameters rather than the raw data from edge devices, FL significantly enhances data privacy during the cooperative model training process \cite{HFL_NOMA_BBW}.
However, conventional FL culminates in a single unified global model that is distributed across all participating devices.
This approach fails to address the task-specific requirements for devices under diverse tasks, as it does not account for the unique data distributions and specific conditions of each device. 
Consequently, it is unsuitable for tasks requiring personalized solutions.

Federated meta-learning (FML) \cite{FL_ML}, which combines the strengths of FL and meta-learning, offers a promising solution to address this challenge. 
In FML, edge devices cooperatively train a meta-model under the orchestration of an edge server.
This meta-model is then fine-tuned locally at each device using one or few gradient steps, allowing the model to meet specific task requirements.
This mechanism not only preserves data privacy by avoiding the exchange of raw data, but also facilitates customized local model training to meet individual performance requirements.
As a result, FML is better suited to handle the diverse requirements and data distributions of individual devices, overcoming the limitations of traditional FL.

Due to its promising benefits, FML has been extensively studied in prior works, focusing on areas such as algorithm design \cite{FML_init, FML_1, FML_ADMM} and its deployment in wireless systems \cite{FML_JSAC, FML_Com, FML_Letter, FML_blockchain}.
The FML method was first introduced in \cite{FML_init} by integrating the model-agnostic meta-learning (MAML) algorithm into the FL framework. 
This was later expanded upon in \cite{FML_1} with a more in-depth analysis aimed at enhancing personalized performance.
In contrast to the aforementioned gradient descent-based methods, \cite{FML_ADMM} developed an alternating direction method of multipliers (ADMM)-based FML algorithm in non-convex cases.
However, due to the random device scheduling, these FML algorithms often face challenges such as low communication efficiency and slow convergence. 
Additionally, when deploying FML in wireless systems, the need for effective resource allocation becomes critical due to the inherent limitations of system resources.
To address these challenges, the authors in \cite{FML_JSAC} jointly optimized device scheduling and resource allocation by considering both the devices' contribution to FML performance and the associated training time and energy costs.
In \cite{FML_Com}, a novel refined FML algorithm was introduced to reduce communication overhead during the training process, while \cite{FML_Letter} proposed a distance-based weighted model aggregation mechanism to accelerate FML convergence.
Furthermore, \cite{FML_blockchain} combined blockchain technology and game theory to design an incentive mechanism for device scheduling, enhancing FML efficacy.
Nevertheless, the aforementioned works primarily focus on optimizing the overall performance of FML in wireless systems, without adequately addressing the individual task requirements of devices with diverse needs.

Motivated by these observations, this paper proposes an FML framework over non-orthogonal multiple access (NOMA) networks to enhance the communication efficiency.
The concept of the value of learning (VoL) is introduced as a novel metric to capture the individual requirements of devices in FML training, taking into account both the desired local model accuracy and the associated total time and energy costs.
Besides, we propose the task level weight (TLW) to measure the importance of different tasks, which incorporates a task requirements-related factor and a fairness-related factor.
To maximize the TLW-based VoL across all devices, a non-convex mixed-integer non-linear programming (MINLP) problem is formulated, aiming to jointly optimize device scheduling and resource allocation.
Since the problem involves both discrete and continuous variables, a parameterized deep Q-network (PDQN)-based deep reinforcement learning method is developed to solve it.
Simulations are conducted to verify the performance of the proposed schemes.

\section{System Model and Problem Formulation}

\begin{figure}
	\centering
	\includegraphics[width=3in]{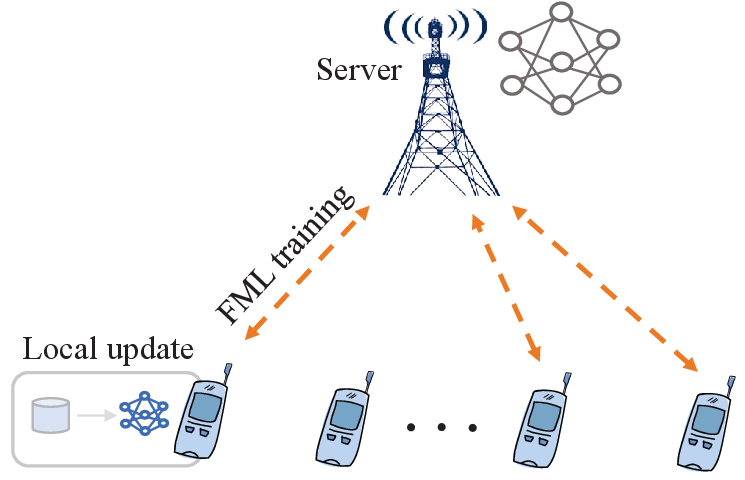}
	\caption{Federated meta-learning system model.}
	\label{SystemModel}
\end{figure}

As shown in Fig. \ref{SystemModel}, a FML system is considered, where an edge server aids in model training for a set ${\cal N}$ of $N$ devices designated for diverse tasks.
Unlike traditional FL, which distributes a common global model to all clients, FML aims to collaboratively train a meta-model utilizing data distributed among devices. 
This meta-model can then be fine-tuned for specific tasks on each device through a few gradient descent steps.
The details of the FML process are described as follows.

\subsection{FML Training}
Each device $n \in {\cal N}$ owns a labeled dataset ${{\cal{D}} _n} = \left\{ {\left( {{{\bf{x}}_i},{y_i}} \right)} \right\}_{i = 1}^{ {{D_n}} }$, where ${D_n}$ is the number of data samples, ${{\bf{x}}_i}$ and ${y_i}$ denote the $i$-th data sample and its label, respectively.
Define $\ell \left( {{\bf{\omega }};x,y} \right)$ as the loss function of parameter ${\bf{\omega }} \in \mathbb{R}^d $ for device $n$. 
The objective of FML is to minimize the average of the meta-function ${L_n}\left( {\bf{\omega }} \right)$ across all devices, which can be expressed as 
\begin{equation}\label{}
	\begin{aligned}
		\mathop {\min }\limits_{\bf{\omega }} \frac{1}{N}\sum\limits_{n \in {\cal N}} {{L_n}\left( {\bf{\omega }} \right)} = \frac{1}{N}\sum\limits_{n \in {\cal N}}{l_n}\left( {{\bf{\omega }} - \alpha \nabla {l_n}\left( {\bf{\omega }} \right)} \right),
	\end{aligned}
\end{equation}
where ${{l_n}\left( {\bf{\omega }} \right) = {\mathbb{E}_{(x,y)}[\ell \left( {{\bf{\omega }};x,y} \right)]}}$ denotes the expected loss function of device $n$ over its data distribution, and $\alpha$ is the learning rate.

In the $t$-th global round, device $n$ conducts several steps of stochastic gradient descent (SGD) to update the local model based on its meta-function ${L_n}\left( {\bf{\omega }} \right)$. 
Specifically, at the $k$-th step of SGD, the local model of device $n$ is updated as 
\begin{equation}\label{}
	\begin{aligned}
		{\bf{\omega }}_n^{t,k} = {\bf{\omega }}_n^{t,k - 1} - \beta \nabla {L_n}\left( {{\bf{\omega }}_n^{t,k - 1}} \right),
	\end{aligned}
\end{equation}
where $\beta$ denotes the meta-learning rate and the gradient of meta-function, i.e., $\nabla {L_n}\left( {{\bf{\omega }}} \right)$, is written as
\begin{equation}\label{grad}
	\begin{aligned}
		\nabla {L_n}\left( {\bf{\omega }} \right) = \left( {{{I}} - \alpha {\nabla ^2}{l_n}\left( {\bf{\omega }} \right)} \right)\nabla {l_n}\left( {{\bf{\omega }} - \alpha \nabla {l_n}\left( {\bf{\omega }} \right)} \right).
	\end{aligned}
\end{equation}
It is computationally costly to compute the gradient $\nabla {l_n}\left( {{\bf{\omega }}} \right)$ at each round.
Thus, $\nabla {l_n}\left( {{\bf{\omega }}} \right)$ and ${\nabla ^2}{l_n}\left( {\bf{\omega }} \right)$ are substituted using their unbiased estimates $\tilde\nabla {l_n}\left( {{\bf{\omega }}} \right)$ and ${\tilde\nabla ^2}{l_n}\left( {\bf{\omega }} \right)$ for any data batch $\tilde D_n$ \cite{FML_3}, which are calculated as 
\begin{equation}\label{}
	\begin{aligned}
		\tilde \nabla {l_n}\left( {{\bf{\omega }}, \tilde D_n} \right) = \frac{1}{{\left| {\tilde D_n} \right|}}\sum\limits_{\left( {x,y} \right) \in {\cal D}_n} {\nabla \ell \left( {{\bf{\omega }};x,y} \right)}, 
	\end{aligned}
\end{equation}
\begin{equation}\label{}
	\begin{aligned}
		{\tilde \nabla ^2}{l_n}\left( {{\bf{\omega }}, \tilde D_n} \right) = \frac{1}{{\left| {\tilde D_n} \right|}}\sum\limits_{\left( {x,y} \right) \in {\cal D}_n} {{\nabla ^2}\ell \left( {{\bf{\omega }};x,y} \right)}.
	\end{aligned}
\end{equation}
Given $\tilde\nabla {l_n}\left( {{\bf{\omega }}} \right)$ and ${\tilde\nabla ^2}{l_n}\left( {\bf{\omega }} \right)$, the estimated meta-function gradient $ \tilde\nabla{L_n}\left( {\bf{\omega } }\right)$ can be given by 
\begin{equation}\label{grad1}
	\begin{aligned}
		\tilde\nabla {L_n}\left( {\bf{\omega }} \right) =& \left( {{{I}} - \alpha {\tilde\nabla ^2}{l_n}\left( {{\bf{\omega }}, \tilde D_n''} \right)} \right) \\
		& \times \tilde\nabla {l_n}\left( {{\bf{\omega }} - \alpha\tilde \nabla {l_n}\left( {\bf{\omega }}, \tilde D_n\right)},  \tilde D_n'\right),
	\end{aligned}
\end{equation}
where $\tilde D_n$, $\tilde D_n'$ and $\tilde D_n''$ are independent data batches \cite{FML_JSAC}.

Subsequently, edge devices upload their updated local model parameters to the server via wireless networks.
The global model is updated at the server in an average manner, i.e.,
\begin{equation}\label{}
	\begin{aligned}
		{{\bf{\omega }}^{t + 1}} = \frac{1}{N_s}\sum\limits_{n \in {\cal N}_s} {{\bf{\omega }}_n^t},
	\end{aligned}
\end{equation}
where ${\cal N}_s$ represents the set of participating devices at round $t$, and ${N_s}$ denotes the corresponding size.
In the next global round, the server broadcasts the updated meta-model to all devices, and the above FML training process repeats until convergence is achieved.
Note that we reasonably neglected the downlink transmission in FML, due to the server's significantly higher transmission power compared to edge devices \cite{downlink}.

It can be obviously observed that the major difference between FL and FML lies in the local update phase. 
In FML, local models are fine-tuned to cater for specific tasks, so that the task-oriented learning can be achieved.

\subsection{Computation and Communication Models} 
For each edge device $n$, we define $f_n$ and $c_n$ as the CPU cycle frequency and required cycles for one data sample, respectively.
For simplicity of analysis, we consider the one-step local update at each device $n$ in this paper, and its total required number of CPU cycles is denoted by ${c_n}{D_n}$.
Thus, the computational time of device $n$ is calculated as 
\begin{equation}\label{}
	\begin{aligned}
		T_n^{\text {cmp}} = \frac{{{c_n}{D_n}}}{{{f_n}}}.
	\end{aligned}
\end{equation}
The corresponding energy consumption for local model update can be given as 
\begin{equation}\label{}
	\begin{aligned}
		E_n^{\text {cmp}} = \frac{\tau }{2}{c_n}{D_n}f_n^2,
	\end{aligned}
\end{equation}
where ${\tau / 2}$ denotes the effective capacitance coefficient of computing chipset \cite{LocComEn}.

NOMA is adopted for transmitting local model parameters, enhancing communication efficiency between devices and the server.
We assume that perfect successive interference cancellation (SIC) can be realized at the receiver \cite{Ding_Survey_2017}.
Let $p_n$ represents the transmitting power of device $n$, and $h_n$ denote the channel gain between the server and device $n$.
The simple decoding order is considered for SIC, following the order of the devices' channel gains, i.e., ${\left| {{h_{1}}} \right|^2}  \ge {\left| {{h_{2}}} \right|^2} \ge  \cdots  \ge {\left| {{h_{{N}}}} \right|^2}$.
Hence, the achievable data rate of device $n$ can be expressed as 
\begin{equation}\label{}
	\begin{aligned}
		{R_n} = {B}{\log _2}\left( {1 + \frac{{{p_n}{{\left| {{h_n}} \right|}^2}}}{{\sum\limits_{k = n + 1}^N {{p_k}{{\left| {{h_k}} \right|}^2}}  + \sigma^2}}} \right),
	\end{aligned}
\end{equation}
where $B$ is the bandwidth, and ${\sigma ^2}$ denotes the variance of additive white Gaussian noise (AWGN).
Define the size of local model parameters as $d_n$, which is assumed to be consistent across all devices due to the fixed dimension of model parameters.
The transmission time and energy consumption of device $n$ are respectively given by
\begin{equation}\label{}
	\begin{aligned}
		T_n^{\text {com}} = \frac{{{d_n}}}{{{R_n}}},
	\end{aligned}
\end{equation}
\begin{equation}\label{}
	\begin{aligned}
		E_n^{\text {com}} = {p_n}{T_n^{\text {com}}}.
	\end{aligned}
\end{equation}

\subsection{Value of Learning}
To capture the learning performance of individual devices with diverse tasks, we introduce the value of learning (VoL) as a metric that quantifies the specific learning requirements of each device, incorporating both positive and negative factors.
Specifically, the positive factor represents the achieved local model accuracy for each device, while the negative factor accounts for the time and energy consumed during model training.

The achieved local model accuracy of device $n$ at the $t$-th round is defined as 
\begin{equation}\label{}
	\begin{aligned}
		{A_n} = \frac{1}{{{{\hat D}_n}}}\sum\limits_{i = 1}^{{{\hat D}_n}} {{{\mathbbm{1}}_{{y_i}}}\left\{ {\xi \left( {{\bf{\omega }}_n^t,{{\bf{x}}_i}} \right)} \right\}},
	\end{aligned}
\end{equation}
where ${{\hat D}_n}$ is the size of the local test dataset of device $n$, ${\mathbbm{1}}\left(  \cdot  \right) \in \left\{0, 1\right\}$ is an indicator function, and ${\mathbbm{1}}\left(  \cdot  \right) = 1$ if and only if the predicted label ${\xi \left( {{\bf{\omega }}_n^t,{{\bf{x}}_i}} \right)}$ is equal to the true label $y_i$.
Let $A_n^{\text {req}}$ denote the required accuracy for the specific task of device $n$.
The positive factor of VoL for device $n$ can be defined as follows:
\begin{equation}\label{}
	\begin{aligned}
		V_n^A = \left\{ \begin{array}{l}
			\frac{A_n}{{A_n^{\text {req} }}}, \ {\text {if}} \ {A_n} \le {{A_n^{\text {req} }}},\\
			1, \ {\text {if}} \ {A_n} > {{A_n^{\text {req} }}}.
		\end{array} \right.
	\end{aligned}
\end{equation}
The definition can be explained as follows: if a device's required local model accuracy is met, it achieves the maximum positive VoL of 1. Otherwise, the positive VoL is a fraction of the achieved local model accuracy relative to the required accuracy.

To describe the negative factors of VoL for each device $n$, we first define the maximum tolerable time and energy consumption as $T_n^{\text {max}}$ and $E_n^{\text {max}}$, respectively. 
Note that the total time and energy consumed by each device during model training must not exceed its specified maximum limits.
Accordingly, the negative factor of VoL related to time for device $n$ can be expressed as a fraction as follows: 
\begin{equation}\label{}
	\begin{aligned}
		V_n^T = \frac{T}{{T_n^{\text {max} }}},
	\end{aligned}
\end{equation}
where $T = {{\mathop {\max }\limits_{n \in {\cal N}} \left\{ {T_n^{\text {cmp}} + T_n^{\text {com}}} \right\}}}$ represents the total time consumed for model training in one global FML round.
The use of the max function indicates that the server employs a synchronous model aggregation mechanism.
Similarly, the negative factor of VoL related to energy consumption for device $n$ is represented as 
\begin{equation}\label{}
	\begin{aligned}
		V_n^E = \frac{E_n}{{E_n^{\text {max} }}},
	\end{aligned}
\end{equation}
where ${E_n} = {E_n^{\text {cmp}} + E_n^{\text {com}}} $ is the total energy consumption of device $n$ in a global round.

\subsection{Task Level Weight}
To evaluate the importance of different tasks, the task level weight (TLW) is introduced to prioritize edge devices based on their specific requirements.
Specifically, we assume that the TLW of device $n$ is influenced by two factors: a requirement-related factor and a fairness-related factor.
For the former, it is intuitively to assume that tasks with larger maximum time and energy consumption constraints indicate a lower importance level.
Additionally, devices with tasks requiring higher accuracy should be assigned greater weights.
Thus, combining these task requirements of device $n$, the cost-related factor of TLW can be defined as
\begin{equation}\label{}
	\begin{aligned}
		\varepsilon _n^{\text {req} } = \frac{1}{{{\lambda _1}T_n^{\text {max} } + {\lambda _2}E_n^{\text {max} } - {\lambda _3}A_n^{\text {req} }}},
	\end{aligned}
\end{equation}
where ${\lambda _1}$, ${\lambda _2}$ and ${\lambda _3}$ are parameters to balance the contributions of time, energy consumption and required model accuracy in determining the task's importance level.

In order to determine the fairness-related factor, we first introduce the concept of age of update (AoU) for local models within FML systems.
Define $z_n^t$ as the aggregation indicator of device $n$ at the $t$-th global round, i.e., if the server schedules device $n$ to upload its trained local model parameter for global aggregation, $z_n^t = 1$; otherwise $z_n^t = 0$.
Thus, the AoU of device $n$ at round $t$ can be given as
\begin{equation}\label{}
	\begin{aligned}
		a_n^t = \left\{ \begin{array}{l}
			a_n^{t - 1} + 1, \ {\text {if}} \ z_n^t = 0,\\
			1, \ {\text {if}} \ z_n^t = 1.
		\end{array} \right.
	\end{aligned}
\end{equation}
Note that a higher AoU value indicates a more outdated local model update, which degrades the performance of FML \cite{AoI}.
Thus, the AoU value across the system should be maintained at a low level, which also ensures fair device participation in the FML system.
We define the fairness-related factor as follows:
\begin{equation}\label{}
	\begin{aligned}
		\varepsilon _n^{{\text {fair}}} = \frac{{a_n^t}}{{\sum\limits_{i \in {\cal N}} {a_i^t} }}.
	\end{aligned}
\end{equation}

Combining the above two defined factors, the TLW of device $n$ is expressed as 
\begin{equation}\label{}
	\begin{aligned}
		\varepsilon _n = \varepsilon _n^{\text {req} } + \varepsilon _n^{{\text {fair}}}.
	\end{aligned}
\end{equation}

\subsection{Problem Formulation}
In this paper, we consider the maximization problem of TLW-based VoL for all devices in the FML system, which is formulated as
\begin{subequations}\label{OP}
	\begin{align}
		\mathop {\max }\limits_{{\bf{z}},{\bf{p}},{\bf{f}}} \quad & \sum\limits_{n \in {\cal N}} {{\varepsilon _n}{z_n}\left( {{\eta _1}V_n^A - {\eta _2}V_n^T - {\eta _3}V_n^E} \right)}  \label{OP_fun} \\
		\st\ \quad  & {z_n} \in \left\{ {0,1} \right\}, \forall n \in {\cal N}, \label{OP_con1}\\
		& 0  \le  {p_n} \le p_n^{\text {max} }, \forall n \in {\cal N}, \label{OP_con2}\\
		& 0  \le {f_n} \le f_n^{\text {max} }, \forall n \in {\cal N}, \label{OP_con3}
	\end{align}
\end{subequations}
where ${\eta _1}$, ${\eta _2}$ and ${\eta _3}$ are weighing parameters to achieve trade-offs among $V_n^A$, $V_n^T$ and $V_n^E$, which are determined by specific scenarios.
Constraint \eqref{OP_con1} indicates the variable ${z_n}$ is binary.
Constraints \eqref{OP_con2} and \eqref{OP_con3} present the feasible regions of transmitting power and computing frequency for devices, respectively.

Obviously, solving problem \eqref{OP} is challenging due to its mixed-integer and non-convex nature, making conventional optimization techniques unsuitable. Therefore, the deep reinforcement learning method is employed in the following section to effectively address the problem.

\section{PDQN-based Device Scheduling and Resource Allocation Design}
In this section, we first reformulate problem \eqref{OP} as a Markov decision process (MDP) model.
Subsequently, we propose a parameterized deep Q-network (PDQN) algorithm to solve it, accounting for the hybrid discrete and continuous action space.

\subsection{MDP Model}
The basic components of MDP are denoted by $\left\{{{\cal S}, {\cal A}, r, {\cal P}} \right\}$, where ${\cal S}$ represents the agent's state space, ${\cal A}$ denotes the agent's possible action space given ${\cal S}$, $r$ is the immediate reward by interacting with the environment, and ${\cal P}$ denotes the probability of state transition.
The details of the MDP model are provided below.

\subsubsection{\bf State space}
We define the agent's action space from two aspects: the instantaneous channel information ${h_{n}}$ and the TLW ${\varepsilon _n}$ of each device.
The former reflects the dynamic wireless communication environment, while the latter represents the priority of devices during FML training.
Hence, the action space of agent at time slot $j$ is denoted as ${S_j} = \left\{ {h_{n}^j, \varepsilon _n, \forall n \in {{\cal N}}} \right\} \in {\cal S}$.

\subsubsection{\bf Action space}
We incorporate the optimization variables of problem \eqref{OP} into the action space, including the discrete variable $z_n$ and continuous variables $p_n$ and $f_n$.
Thus, given the state information at time slot $j$, the action of agent is denoted by ${A_j} = \left\{ {z_n^j, p_n^j,f_n^j, \forall n \in {{\cal N}}} \right\} \in {\cal A}$.

\subsubsection{\bf Reward}
At time slot $j$, the agent interacts with the environment by taking action $A_j$, contributing to an achievable reward $r_j$ and a transition to a new state $S_{j+1}$ in the next time slot.
We define the objective function in problem \eqref{OP} as the reward, which aims to maximize the TLW-based VoL for all devices.
The reward function of agent at time slot $j$ is given by
\begin{equation}\label{}
	\begin{aligned}
		r_j = \left\{ \begin{array}{l}
			V_{\text {total}}^j, \ {\text {if}} \ V_{\text {total}}^j > 0,\\
			0, \ {\text {if}} \ V_{\text {total}}^j \le 0,
		\end{array} \right.
	\end{aligned}
\end{equation}
where $V_{\text {total}}^j $ represents the objective function in \eqref{OP} at time slot $j$.

\begin{figure}[t]
	\centering
	\includegraphics[width=0.5\textwidth]{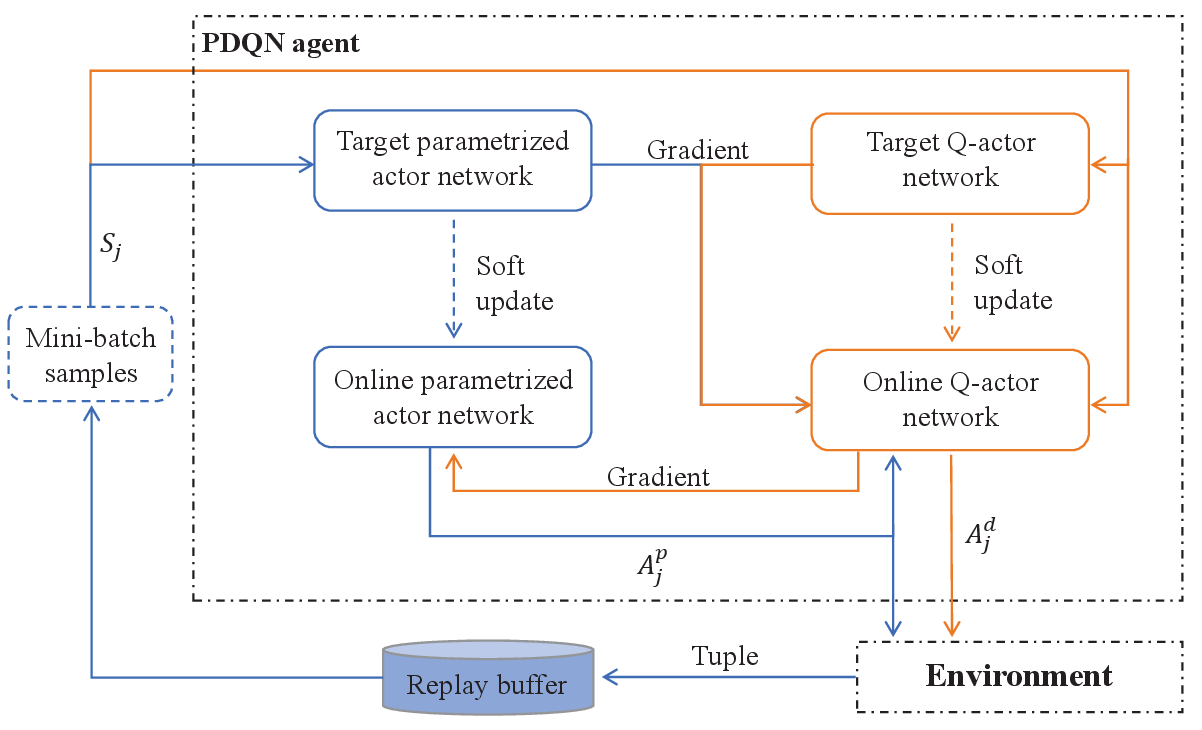}
	\caption{Training framework of PDQN.}
	\label{PDQN}
\end{figure}

\subsection{PDQN-based Algorithm}
Since the formulated MDP model involves hybrid discrete and continuous actions, common DRL algorithms like deep-Q learning (DQN) and deep deterministic policy gradient (DDPG) are unsuitable, as they only handle either discrete or continuous action spaces.
In this case, PDQN method is a promising solution for addressing hybrid action spaces, as it combines a Q-actor network and a parameterized actor network, as shown in Fig. \ref{PDQN}.
Additionally, both networks include a target network and an online network to mitigate the overestimation issue.
Define $\nu \left( {S_j|{{\bf{\theta }} }} \right)$ as the online parameterized actor network, where $\theta$ represents the corresponding parameters.
Hence, its output, i.e., the parameterized actions $A_j^p$, can be expressed as 
\begin{equation}\label{para_action}
	\begin{aligned}
	  A_j^p = \nu \left( {S_j|{{\bf{\theta }} }} \right) + noise,
	\end{aligned}
\end{equation}
where $noise$ is the added Gaussian noise for action exploration.
The online Q-actor network serves for the calculation of state-action value function $Q\left( {S, A^p, A^d} \right)$.
It is utilized to select the appropriate discrete actions $A^d$ by evaluating the parameterized actions.
Following the concept of DQN, the discrete actions that maximize the Q-value are selected, which can be expressed as
\begin{equation}\label{dis_action}
	\begin{aligned}
		A_j^d = \mathop {\arg \max }\limits_{{{\cal A}^d}} Q\left( {{S_j},A_j^p,A_j^d{\rm{|}}\varphi } \right),
	\end{aligned}
\end{equation}
where $\varphi$ denotes the online Q-actor network's parameters.

\begin{algorithm}[t]
	\footnotesize
	\caption{PDQN-based Algorithm for Joint Device Scheduling and Resource Allocation.}
	\label{algorithm}
	\begin{algorithmic}[1] 
		\STATE \textbf{Initialization:} Online network parameters $\theta$, $\varphi$; target network parameters $\tilde \theta$, $\tilde \varphi$; reply buffer maximum capacity $\cal G$.
		\STATE \textbf{for} each episode \textbf{do}
		\STATE  \qquad Initialize random $noise$ for action exploration;
		\STATE  \qquad Observed the initial state $S_0$ from environment;
		\STATE  \qquad \textbf{for} each time slot \textbf{do}
		\STATE  \qquad \qquad Select parameterized actions $A_j^p$ and discrete actions $A_j^d$ based \\ \qquad \qquad on \eqref{para_action} and \eqref{dis_action}, respectively; 
		\STATE  \qquad \qquad Execute actions $A_j^p$ and $A_j^d$, obtain reward and transit to next \\ \qquad \qquad state $S_{j+1}$;
		\STATE  \qquad \qquad Store tuple $\left({{S_j},{A_j^p},{A_j^d},{r_j},{S_{j + 1}}} \right)$ into the replay buffer;
		\STATE  \qquad \qquad \textbf{if} reach $\cal G$ \textbf{then}
		\STATE  \qquad \qquad \qquad Randomly sample a mini-batch data $\cal M$;
		\STATE  \qquad \qquad \qquad Update online parameterized actor network and Q-actor \\ \qquad \qquad \qquad network according to \eqref{para_network} and \eqref{q_network}; 
		\STATE  \qquad \qquad \qquad Update target networks using \eqref{target_network}.
		\STATE  \qquad \qquad \textbf{end if}
		\STATE  \qquad \textbf{end for}
		\STATE \textbf{end for}
	\end{algorithmic}
\end{algorithm}

By executing the continuous and discrete actions at time slot $j$, the agent interacts with the environment, obtaining the instantaneous reward $r_j$ and transitioning to the next state $S_{j+1}$.
Using the experience replay mechanism, the tuple $\left( {{S_j},{A_j^p},{A_j^d},{r_j},{S_{j + 1}}} \right)$ is stored at the experience buffer with maximum capacity $\cal G$.
Once reaching the maximum buffer limit, a mini-batch of samples $\cal M$ is randomly selected from the buffer to update networks.
We update the online parameterized actor network via the policy gradient method as follows \cite{PDQN}:
\begin{equation}\label{para_network}
	\begin{aligned}
		{\nabla _\theta }L\left( \theta  \right) = \mathop {\mathbb{E}}\limits_{\cal M} \left[ {{\nabla _\theta }\nu \left( {S^m{\rm{|}}\theta } \right) \times {\nabla _{{A^{p,m}}}}Q\left( {S^m,{A^{p,m}},{A^{d,m}}{\rm{|}}\varphi } \right)} \right],
	\end{aligned}
\end{equation}
where $L\left( \theta  \right)$ is the loss function with respect to $\theta $, $m$ indicates the index for mini-batch sample, and ${\mathbb{E}}\left(  \cdot  \right)$ denotes the expectation operator.
The online Q-actor network is updated through minimizing the following loss function $L\left( \varphi  \right)$:
\begin{equation}\label{q_network}
	\begin{aligned}
		L\left( \varphi  \right) = \mathop {\mathbb{E}}\limits_{\cal M} \left[ {\left( {{y^m} - Q\left( {{S^m},{A^{p,m}},{A^{d,m}}{\rm{|}}\varphi } \right)} \right)^2} \right],
	\end{aligned}
\end{equation}
where $y^m = {r^m} + \kappa \max \tilde Q\left( {{S^m},\tilde \nu \left( {{{\tilde S}^m}{\rm{|}}\tilde \theta } \right),{A^{d,m}}{\rm{|}}\tilde \varphi } \right)$ represents the current target Q-value, $\kappa$ is the discount factor, and $\tilde \theta$ and $\tilde \varphi$ represent the parameters of  target parameterized actor network $\tilde \nu \left( {S|{{{\tilde \theta }} }} \right)$ and target Q-actor network $\tilde Q\left( {{S},A^p,A^d{\rm{|}}\tilde \varphi } \right)$, respectively.
They are updated using the following soft update mechanism:
\begin{equation}\label{target_network}
	\begin{aligned}
		& {{{\tilde \theta }}} \leftarrow \zeta {{{\theta }} } + \left( {1 - \zeta } \right){{{ \tilde \theta }}},\\
		& {{{\tilde \varphi }}} \leftarrow \zeta {{{\varphi }}} + \left( {1 - \zeta } \right){{{\tilde \varphi }}},
	\end{aligned}
\end{equation}
where $\zeta$ indicates the soft update parameter.
The PDQN-based algorithm for addressing problem \eqref{OP} is summarized in Algorithm \ref{algorithm}.

\section{Simulation Results}
In this section, numerical simulations are conducted to evaluate the performance of the proposed schemes in FML systems.
We consider a square area with a side length of $500$ meters, where the edge server is located at the center and $10$ edge devices randomly distributed throughout the area.
The CIFAR-10 data is distributed among devices in a non-independent and identically distributed (non-IID) fashion, targeting diverse tasks with specific requirements.
We set the required local model accuracy ${{A_n^{\text {req} }}}$ for each device as a random value between $0.7$ to $1.0$.
Additionally, the maximum time and energy consumption for each device, $T_n^{\text {max}}$ and $E_n^{\text {max}}$, are randomly chosen from the ranges of $0.1$ to $10$ seconds and $0.01$ to $1$ joules, respectively.
The other parameters are presented in Table \ref{Para}. 

\begin{table}[t]
	\footnotesize
	\begin{center}
		\caption{\protect\\\textsc{Simulation Parameters}}\vspace{+1em}
		\label{Para}
		\begin{tabular}{c|c}
			\hline
			Parameter & Value\\  \hline
			Carrier frequency & $1$ GHz\\
			Bandwidth, $B $ & $1$ MHz \\
			Path loss exponent & $3.76$\\
			AWGN spectral density & $-174$ dBm/Hz\\
			Maximum transmit power, $p_n^{\text{max}}$ & $0.1 $  W\\
			CPU cycles for each sample, $c_n$ & $10^7$ \\
			Maximum computation frequency, $f_n^{\text{max}}$ & $10$ GHz\\
			Local model size, $d_n $ & $ 1$ Mbit\\
			\hline
		\end{tabular}
	\end{center}
\end{table}

\begin{figure}[t]
	\centering
	\includegraphics[width=0.45\textwidth]{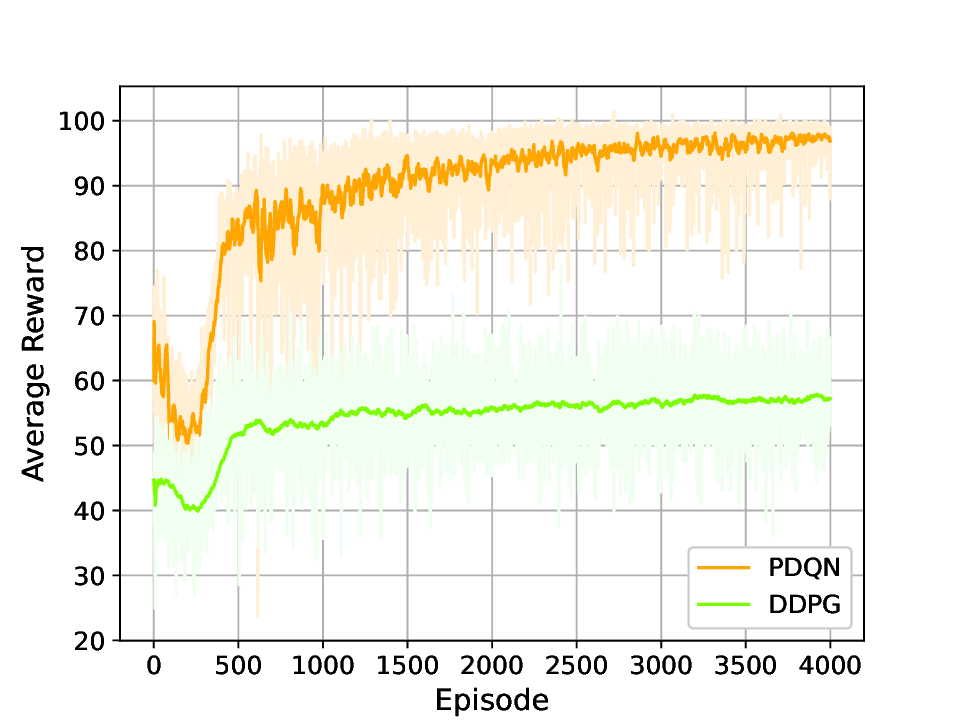}
	\caption{Convergence of PDQN and DDPG algorithms.}
	\label{PD_DD}
\end{figure}

Fig. \ref{PD_DD} illustrates the convergence performance of the PDQN and DDPG algorithm across training episodes.
The figures in light colors represent the instant reward obtained at each episode, while the figures in dark colors indicate the average reward of $20$ episodes.
As shown, the PDQN-based algorithm achieves a higher reward value compared to the DDPG-based algorithm upon convergence.
This is because the DDPG-based algorithm is limited to handling only continuous actions, and it requires rounding to approximate the discrete actions. 
This rounding introduces inaccuracies, leading to an inevitable reduction in performance compared to the PDQN-based algorithm, which is specifically designed to handle both discrete and continuous actions effectively.

\begin{figure}[t]
	\centering
	\includegraphics[width=0.45\textwidth]{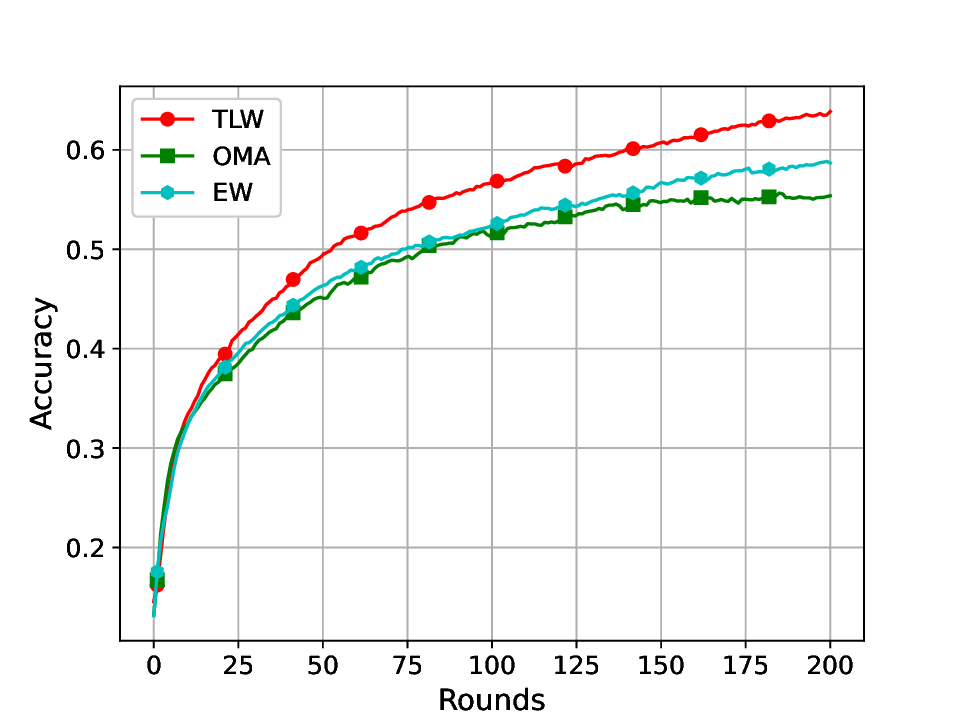}
	\caption{FML performance on non-IID CIFAR-10 dataset.}
	\label{FML}
\end{figure}

In Fig. \ref{FML}, we compare the proposed TLW-based scheme with two benchmark schemes, namely the orthogonal multiple access (OMA) scheme and the equal weight (EW) scheme, in terms of FML performance.
In the case of the OMA scheme, OMA is utilized for transmitting model parameters between devices and the server.
For the EW scheme, the devices' tasks are regarded as having the same level of importance, leading to a rotation approach for device scheduling.
As observed from Fig. \ref{FML}, the proposed TLW-based scheme outperforms the two benchmarks in test accuracy performance, as it jointly accounts for the specific requirements of each device and the fairness factor during the FML training process.
However, the FML performance in the OMA and EW schemes is limited due to inferior communication efficiency and a lack of device priority, respectively.

\begin{figure}[t]
	\centering
	\includegraphics[width=0.45\textwidth]{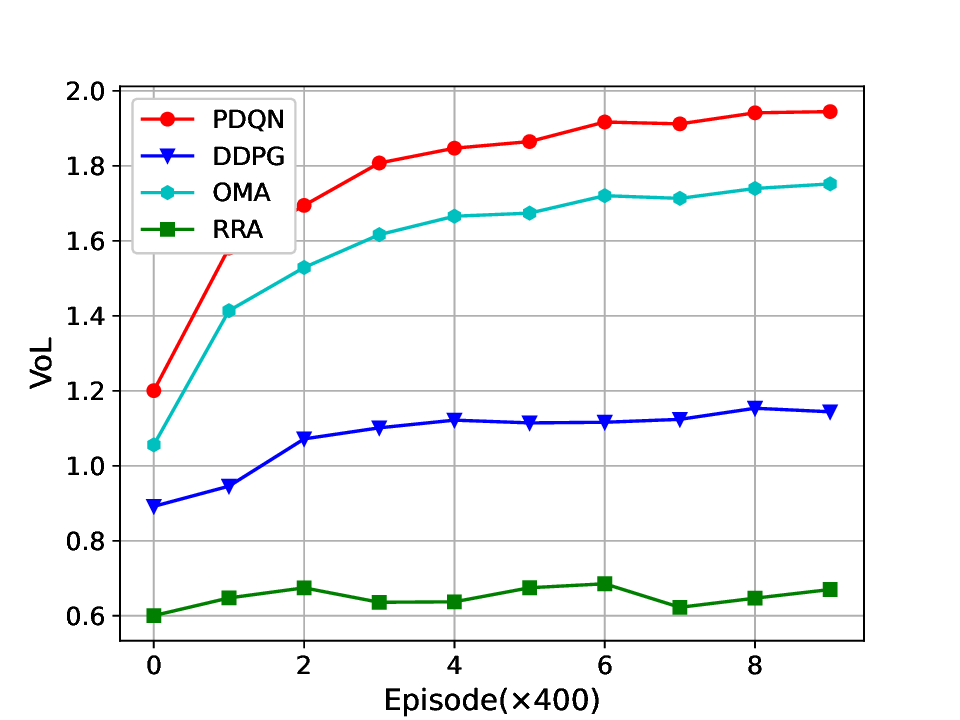}
	\caption{VoL performance versus episode.}
	\label{VoL}
\end{figure}

Fig. \ref{VoL} illustrates the VoL performance of the proposed PDQN-based algorithm in comparison with the DDPG-based algorithm, the OMA scheme, and the random resource allocation (RRA) scheme.
It is observed that the proposed PDQN-based algorithm over NOMA networks achieves the highest VoL with increasing training episodes, followed by the OMA scheme. This can be attributed to the fact that the OMA scheme also employs the PDQN approach for the joint optimization of device scheduling and resource allocation, effectively addressing the hybrid discrete and continuous actions.
Nevertheless, the DDPG-based algorithm struggles to effectively manage discrete variables, which leads to a notable reduction in VoL performance.
There is no doubt that the RRA scheme has the worst VoL performance, due to its random nature in resource allocation.
Hence, the proposed PQDN-based algorithm is effective in enhancing VoL performance in the considered FML system.

\section{Conclusion}
In this paper, we proposed an FML system over NOMA networks, where NOMA improves communication efficiency for transmitting local model parameters between edge devices and the server.
The VoL was introduced as a novel metric to capture the diverse individual requirements of devices during FML training, incorporating both the positive factor of desired local model accuracy and the negative factor of consumed costs.
Additionally, the TLW was utilized to measure the importance of devices' tasks, based on two factors: task requirements and fairness.
We formulated a maximization problem for the sum of TLW-based VoL across all devices, which was effectively addressed via the PQDN-based algorithm to handle the hybrid discrete and continuous optimization variables.
Simulation results demonstrated that our proposed scheme outperforms the benchmarks in enhancing FML performance and improving the total VoL of devices.

\bibliographystyle{IEEEtran}
\bibliography{EEref}

\begin{thebibliography}{10}
\providecommand{\url}[1]{#1}
\csname url@samestyle\endcsname
\providecommand{\newblock}{\relax}
\providecommand{\bibinfo}[2]{#2}
\providecommand{\BIBentrySTDinterwordspacing}{\spaceskip=0pt\relax}
\providecommand{\BIBentryALTinterwordstretchfactor}{4}
\providecommand{\BIBentryALTinterwordspacing}{\spaceskip=\fontdimen2\font plus
\BIBentryALTinterwordstretchfactor\fontdimen3\font minus
  \fontdimen4\font\relax}
\providecommand{\BIBforeignlanguage}[2]{{%
\expandafter\ifx\csname l@#1\endcsname\relax
\typeout{** WARNING: IEEEtran.bst: No hyphenation pattern has been}%
\typeout{** loaded for the language `#1'. Using the pattern for}%
\typeout{** the default language instead.}%
\else
\language=\csname l@#1\endcsname
\fi
#2}}
\providecommand{\BIBdecl}{\relax}
\BIBdecl

\bibitem{FL_Mag}
M.~Chen, H.~V. Poor, W.~Saad, and S.~Cui, ``Wireless communications for
  collaborative federated learning,'' \emph{IEEE Commun. Mag.}, vol.~58,
  no.~12, pp. 48--54, 2020.

\bibitem{HFL_NOMA_BBW}
B.~Wu, F.~Fang, X.~Wang, D.~Cai, S.~Fu, and Z.~Ding, ``Client selection and
  cost-efficient joint optimization for {NOMA}-enabled hierarchical federated
  learning,'' \emph{IEEE Trans. Wireless Commun.}, pp. 1--1, 2024.

\bibitem{FL_ML}
X.~Liu, Y.~Deng, A.~Nallanathan, and M.~Bennis, ``Federated learning and meta
  learning: Approaches, applications, and directions,'' \emph{IEEE Commun.
  Surv. Tutorials}, vol.~26, no.~1, pp. 571--618, 2024.

\bibitem{FML_init}
F.~Chen, M.~Luo, Z.~Dong, Z.~Li, and X.~He, ``Federated meta-learning with fast
  convergence and efficient communication,'' \emph{arXiv preprint
  arXiv:1802.07876}, 2018.

\bibitem{FML_1}
Y.~Jiang, J.~Kone{\v{c}}n{\`y}, K.~Rush, and S.~Kannan, ``Improving federated
  learning personalization via model agnostic meta learning,'' \emph{arXiv
  preprint arXiv:1909.12488}, 2019.

\bibitem{FML_ADMM}
S.~Yue, J.~Ren, J.~Xin, S.~Lin, and J.~Zhang, ``Inexact-{ADMM} based federated
  meta-learning for fast and continual edge learning,'' in \emph{Proc. 22nd
  Int. Symp. Theory, Algorithmic Found., Protocol Design Mobile Netw. Mobile
  Comput.}, 2021, pp. 91--100.

\bibitem{FML_JSAC}
S.~Yue, J.~Ren, J.~Xin, D.~Zhang, Y.~Zhang, and W.~Zhuang, ``Efficient
  federated meta-learning over multi-access wireless networks,'' \emph{IEEE J.
  Select. Areas Commun.}, vol.~40, no.~5, pp. 1556--1570, 2022.

\bibitem{FML_Com}
F.~Yu, H.~Lin, X.~Wang, S.~Garg, G.~Kaddoum, S.~Singh, and M.~M. Hassan,
  ``Communication-efficient personalized federated meta-learning in edge
  networks,'' \emph{IEEE Trans. Netw. Serv.}, vol.~20, no.~2, pp. 1558--1571,
  2023.

\bibitem{FML_Letter}
L.~Zhang, C.~Zhang, and B.~Shihada, ``Efficient wireless traffic prediction at
  the edge: A federated meta-learning approach,'' \emph{IEEE Commun. Lett.},
  vol.~26, no.~7, pp. 1573--1577, 2022.

\bibitem{FML_blockchain}
E.~Baccour, A.~Erbad, A.~Mohamed, M.~Hamdi, and M.~Guizani, ``A
  blockchain-based reliable federated meta-learning for metaverse: A dual game
  framework,'' \emph{IEEE Internet Things J.}, vol.~11, no.~12, pp.
  22\,697--22\,715, 2024.

\bibitem{FML_3}
A.~Fallah, A.~Mokhtari, and A.~Ozdaglar, ``Personalized federated learning with
  theoretical guarantees: A model-agnostic meta-learning approach,'' \emph{in
  Proc. NIPS}, vol.~33, pp. 3557--3568, 2020.

\bibitem{downlink}
C.~T. Dinh, N.~H. Tran, M.~N.~H. Nguyen, C.~S. Hong, W.~Bao, A.~Y. Zomaya, and
  V.~Gramoli, ``Federated learning over wireless networks: Convergence analysis
  and resource allocation,'' \emph{{IEEE}/ACM Trans. Netw.}, vol.~29, no.~1,
  pp. 398--409, 2021.

\bibitem{LocComEn}
T.~D. Burd and R.~W. Brodersen, ``Processor design for portable systems,''
  \emph{J. VLSI Sig. Proc. Syst.}, vol.~13, no. 2-3, pp. 203--221, 1996.

\bibitem{Ding_Survey_2017}
Z.~Ding, X.~Lei, G.~K. Karagiannidis, R.~Schober, J.~Yuan, and V.~K. Bhargava,
  ``A survey on non-orthogonal multiple access for {5G} networks: {Research}
  challenges and future trends,'' \emph{IEEE J. Select. Areas Commun.},
  vol.~35, no.~10, pp. 2181--2195, Oct. 2017.

\bibitem{AoI}
W.~Dai, Y.~Zhou, N.~Dong, H.~Zhang, and E.~P. Xing, ``Toward understanding the
  impact of staleness in distributed machine learning,'' \emph{ArXiv}, vol.
  abs/1810.03264, 2018.

\bibitem{PDQN}
N.~Lin, H.~Tang, L.~Zhao, S.~Wan, A.~Hawbani, and M.~Guizani, ``A {PDDQNLP}
  algorithm for energy efficient computation offloading in {UAV}-assisted
  {MEC},'' \emph{IEEE Trans. Wireless Commun.}, vol.~22, no.~12, pp.
  8876--8890, 2023.

\end{thebibliography}

\end{document}